\documentclass[11pt]{article}

\usepackage[]{acl}

\usepackage{booktabs} 
\usepackage{multirow}
\usepackage{times}
\usepackage{latexsym}
\usepackage{ulem}
\usepackage{tcolorbox}
\usepackage{amsmath}
\usepackage{amsfonts}
\usepackage[T1]{fontenc}

\usepackage[utf8]{inputenc}

\usepackage{microtype}

\usepackage{inconsolata}
\usepackage{graphicx}
\definecolor{ack}{RGB}{96,169,23}
\definecolor{rel}{RGB}{27,161,226}
\definecolor{app}{RGB}{255,128,0}
\definecolor{rea}{RGB}{102,0,204}
\definecolor{ans}{RGB}{155,0,0}
\usepackage{url}
\definecolor{MyDarkGreen}{rgb}{0.02,0.6,0.02}
\definecolor{MyDarkRed}{rgb}{0.8,0.02,0.02}
\definecolor{MyDarkOrange}{rgb}{0.40,0.2,0.02}
\definecolor{MyPurple}{RGB}{111,0,255}
\definecolor{MyRed}{rgb}{1.0,0.0,0.0}
\definecolor{MyGold}{rgb}{0.75,0.6,0.12}
\definecolor{MyDarkgray}{rgb}{0.66, 0.66, 0.66}

\title{Reason-KE++: Aligning the Process, Not Just the Outcome,\\ for Faithful LLM Knowledge Editing}

\author{%
  Yuchen Wu$^{1}$,
  Liang Ding$^{2}$\thanks{Correspond to Liang Ding \texttt{liangding.liam@gmail.com}},
  Li Shen$^{3}$,
  Dacheng Tao$^{4}$\\
  $^{1}$Shanghai Jiao Tong University, China 200240\\
  $^{2}$The University of Sydney, Australia 2006\\
  $^{3}$Shenzhen Campus of Sun Yat-sen University, China 518107\\
  $^{4}$Nanyang Technological University, Singapore 639798 
}

\begin{document}
\maketitle
\begin{abstract}
Aligning Large Language Models (LLMs) to be faithful to new knowledge in complex, multi-hop reasoning tasks is a critical, yet unsolved, challenge. We find that SFT-based methods, e.g., Reason-KE~\cite{reasonke}, while state-of-the-art, suffer from a "faithfulness gap": they optimize for format mimicry rather than sound reasoning. This gap enables the LLM's powerful parametric priors to override new contextual facts, resulting in critical factual hallucinations (e.g., incorrectly reasoning "Houston" from "NASA" despite an explicit edit). To solve this core LLM alignment problem, we propose Reason-KE++, an SFT+RL framework that instills process-level faithfulness. Its core is a Stage-aware Reward mechanism that provides dense supervision for intermediate reasoning steps (e.g., Decomposition, Sub-answer Correctness). Crucially, we identify that naive outcome-only RL is a deceptive trap for LLM alignment: it collapses reasoning integrity (e.g., 19.00\% Hop acc) while superficially boosting final accuracy. Our process-aware framework sets a new SOTA of 95.48\% on MQUAKE-CF-3k (+5.28\%), demonstrating that for complex tasks, aligning the reasoning process is essential for building trustworthy LLMs. Our code will be available at: \url{https://github.com/YukinoshitaKaren/Reason-KE}.
\end{abstract}

\section{Introduction}

\begin{figure}[t]
\centering
\includegraphics[width=1\columnwidth]{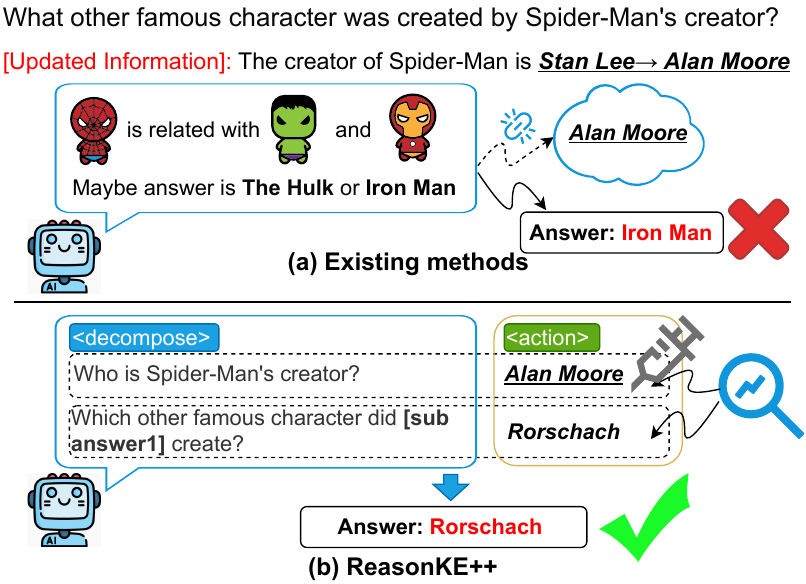}
\vspace{-5pt}
  \caption{
  \textbf{An illustration of our core motivation}. (a) Existing methods often take an unfaithful shortcut based on strong priors, ignoring updated information and leading to incorrect answers. (b) Our ReasonKE++ decomposes the multi-hop query, ensuring a faithful reasoning process that correctly utilizes the new knowledge.
  }
  \vspace{-15pt}
  \label{fig:figure1}
\end{figure}

Large language models (LLMs)~\cite{grattafiori2024llama,qwen25,deepseek} have shown strong capabilities~\cite{survey}, but their fixed parameters struggle with changing world knowledge. Consequently, knowledge editing (KE,~\citet{editing}) has emerged to enable precise modification of specific facts. Current methods are broadly categorized as parameter modification and parameter preservation.

Parameter modification methods~\cite{modifying,rome,memit} can directly change specific parameters to integrate new knowledge. However, current research~\cite{zhang2024uncovering,mquake} doubts that it merely performs surface-level editing without truly understanding. 
In contrast, parameter preservation methods~\cite{EDITCOT,ice} achieve remarkable success by adding extra modules or leveraging the in-context-learning ability of LLMs. Currently, the parameter preservation KE framework performs well in multi-hop question answering (MQA) tasks~\cite{mquake,ripple}, which require models to reason based on updated information. Although in-context learning ability enhances models' understanding of updated knowledge~\cite{IKE}, it can also lead to excessive reliance on facts in the context. So when they encounter noisy or irrelevant knowledge, their performance drops sharply.

Furthermore, existing KE methods neglect the faithfulness of the reasoning process. We find SFT-based method, e.g., ReasonKE~\cite{reasonke}, suffers from a critical "faithfulness gap": they optimize for format mimicry, enabling the LLM's powerful parametric priors to override new contextual facts. This often leads to an unfaithful "shortcut" (see Figure~\ref{fig:figure1}a), where the model ignores the updated knowledge (e.g., "Alan Moore") and defaults to its pre-trained association (e.g., "Stan Lee → Iron Man").

To solve this, we propose \textbf{Reason-KE++}, which ensures a faithful reasoning process by enforcing a structured decomposition of the query (see Figure~\ref{fig:figure1}b). Reason-KE++ is a novel framework designed to fully unleash the model's multi-hop reasoning capabilities while maintaining robustness against distractors.
It tackles problems through a meticulously designed reasoning process, which consists of three steps: 1) \textit{Acknowledge} updated information and the question; 2) \textit{Decompose} the question into sub-questions; and 3) \textit{Act} by sequentially answering these sub-questions to derive the final solution. 
Inspired by Reason-KE, Reason-KE++ explicitly outputs these multiple reasoning steps within a single pass, which circumvents the reliance on complex iterative pipelines.

Specifically, our Reason-KE++ framework involves two phases: 
(1) The first stage focuses on Cold-Start Supervised Fine-Tuning (SFT) to instill initial reasoning patterns in the LLM.
(2) The second stage transitions to reinforcement learning. \textbf{Crucially, we identify that naive, outcome-only RL is a deceptive trap}: our experiments (see Table 6) show it \textit{collapses} reasoning integrity (e.g., 19.00\% Hops acc) while superficially boosting final accuracy.
To solve this, we introduce a novel \textbf{Stage-aware Reward mechanism}. Unlike sparse, outcome-only signals, our method employs a hierarchical reward structure that evaluates both the final answer's correctness and the validity of intermediate reasoning steps. This granular feedback loop discourages shortcut learning and ensures true process-level faithfulness.

We validated the effectiveness across various datasets upon several models. Notably, on the MQuAKE-CF-3k dataset, Reason-KE++ achieved a multi-hop QA accuracy of 95.48\%, marking a significant improvement of 5.28\% over Reason-KE. Moreover, Reason-KE++ exhibits superior reasoning quality by generating more coherent reasoning paths and effectively preventing shortcut learning. The model also demonstrates strong robustness to severe distractions, with its performance declining by only 5.06\% under such conditions.

Our \textbf{contributions} are threefold:
\begin{itemize}
    \item We propose Reason-KE++, a novel two-stage (SFT+RL) framework for knowledge editing that employs a Stage-aware Reward mechanism. Our mechanism decomposes complex reasoning tasks into multiple assessable stages and provides step-by-step supervision, significantly improving the faithfulness of the model's reasoning.
    \item We demonstrate that Reason-KE++ substantially enhances model robustness and mitigates shortcut learning. By explicitly rewarding valid intermediate reasoning steps, our framework trains the model to construct coherent lines of reasoning and ignore irrelevant information, maintaining high performance even in the presence of severe distractors.
    \item We conduct comprehensive experiments on multiple knowledge editing benchmarks, where Reason-KE++ achieves state-of-the-art performance across diverse distractor settings.
\end{itemize}

\section{Preliminary}
\subsection{Knowledge Editing of LLMs.}

The goal of knowledge editing is to efficiently modify specific knowledge encoded within an LLM's parameters~\cite{fast}. A fact is represented as a triplet $f = (s, r, o)$, where $s$ denotes the subject, $r$ the relation, and $o$ the object. The knowledge editing operation updates the object, expressed as $e = (s, r, o \rightarrow o^*)$, for example, \textit{(the United States, the president of \{ \} is, Joe Biden $\rightarrow$ Donald Trump)}. After editing, the model is expected to respond with the updated object “Donald Trump” to a relevant query (e.g., “Who is the president of the United States?”).

\subsection{Multi-hop QA within Knowledge Editing.}

Unlike one-hop questions, answering a multi-hop question $Q$ requires reasoning over a sequence of interdependent facts, or a "chain of facts," $C=[(s_1,r_1,o_1),...,(s_n,r_n,o_n)]$, where $s_{i+1}=o_i$ and $o_n$ is the final answer to $Q$. Under the knowledge editing setting, any alteration to this chain can change the final answer.

Specifically, given a base LLM $p_{\theta}$ and an editing set $ \mathcal{E} = \{e_1,e_2,...,e_m\}$, the task is to produce an edited LLM $p_ {\theta} ^ * $ that can correctly answer a corresponding multi-hop question $Q$. Most previous works~\cite{EDITCOT,mquake,pokemqa} attempt this by finding the "golden path" $ C^*=[(s_1,r_1,o_1),...,(s_i,r_i,o_i^*),...,(s_n^*,r_n,o_n^*)]$ for $Q$ through decomposition and iterative frameworks.

However, this approach faces two fundamental challenges. First, supervising the correctness of intermediate steps is difficult, failing to prevent \textbf{procedural shortcuts}. Second, these methods often overlook the \textbf{noise problem} endemic to real-world scenarios, where the editing set $\mathcal{E}$ may include redundant or irrelevant information. Addressing this gap requires a framework that can ensure step-by-step reasoning faithfulness while simultaneously mitigating the impact of distractors.

\section{Methodology}
\label{sec:methodology}

\begin{figure*}[t]
\centering
\includegraphics[width=1\textwidth]{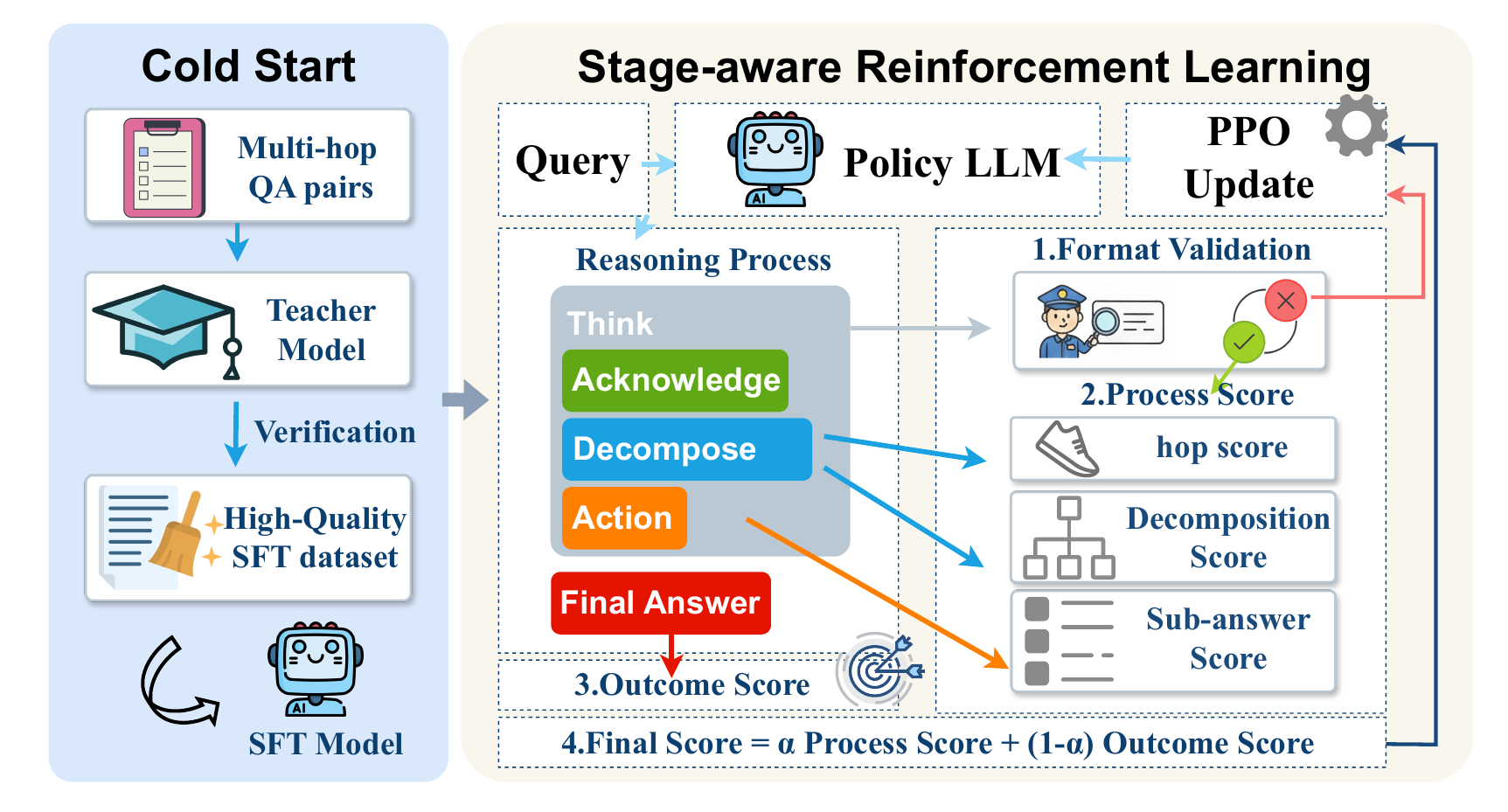}
  \caption{
  \textbf{The two-stage pipeline of ReasonKE++}. It starts with a cold-start SFT phase for foundational learning, followed by a Stage-aware Reinforcement Learning phase. In the RL stage, a dense reward signal, composed of a detailed Process Score and an Outcome Score, is used to optimize the model's ability to generate faithful reasoning.
  }
  \label{fig:method}
  \vspace{-10pt}
\end{figure*}

Reason-KE++ is an RL-based framework. Unlike prior RL approaches, Reason-KE++ decomposes the complex reasoning process into multiple, evaluable stages. By designing a specific reward score for each stage, it transforms the final reward score from sparse to dense. This mechanism guides the model to construct faithful reasoning pathways and effectively mitigates shortcut learning.
As shown in Figure~\ref{fig:method}, Reason-KE++ consists of two stages: a cold-start supervised fine-tuning phase to teach basic reasoning patterns, followed by a reinforcement learning phase with a Stage-aware Reward mechanism.

\subsection{Reasoning Process Design}
To ensure the model's reasoning is both transparent and faithful, we designed a structured process to guide its thinking. The entire thought process is contained within \texttt{<think>}...\texttt{</think>} tags and is organized into three distinct stages, each demarcated by its own special tokens (e.g., \texttt{<acknowledge>}...\texttt{</acknowledge>}). These three stages are:
(1) \textbf{Acknowledge:} the model confirms the updated knowledge and its relevance to the input query.
(2) \textbf{Decompose:} then it breaks the main problem into a series of actionable sub-questions.
(3) \textbf{Act:} it methodically solves each sub-question, explicitly showing the derivation of each intermediate answer highlighted using the \texttt{\textbackslash boxed\{\}}.
Following this detailed thought process, the final answer is delivered, enclosed within \texttt{<answer>} and \texttt{</answer>} tags. 
This structured, machine-parsable format is a necessary prerequisite, as it enables the fine-grained evaluation required by our Stage-aware Reward mechanism in Section~\ref{sec:stage_aware_rl}.

\subsection{Cold Start for Foundational Reasoning}
To prepare for the subsequent reinforcement learning phase, we start with Supervised Fine-Tuning (SFT) to equip the base LLM with the foundational capability to generate our structured reasoning process. To achieve this, we carefully curated a high-quality dataset.

Specifically, our data creation process begins by extracting multi-hop QA pairs from the COUNTERFACT~\cite{easyedit}. We then employ a structured prompt template to guide an advanced teacher model (e.g., GPT-4o-mini) to produce a step-by-step reasoning process for each pair.
Moreover, to ensure the quality of the training data, we apply a strict verification protocol that rectifies formatting errors and discards non-compliant samples. This ensures all outputs have syntactic consistency and structural integrity, making them atomically verifiable for the RL stage. Finally, this SFT process trains the model to acknowledge new facts, decompose multi-hop questions, and logically derive the final answer, establishing a solid foundation for the next stage. More details can be found in Appendix~\ref{sec:Dataset_sample}.

\subsection{Reason-KE++}
\subsubsection{Training Algorithm}

We train Reason-KE++ by employing the Proximal Policy Optimization (PPO) algorithm. The policy $\pi_{\theta}$ is updated by optimizing the PPO objective:
\begin{equation}
\footnotesize
\begin{split}
    \hspace*{-0.6em} \mathcal{J}_\text{PPO}(\theta) = \mathbb{E}{[(q,a)\sim \mathcal{D}, o_{\le t}\sim\pi_{\theta_{\text{old}}}(\cdot\mid q)]} \\
    &  \hspace*{-15em} \left\{ \min \big[ r_t(\theta) \hat{A}_t, \, \text{clip} ( r_t(\theta), 1 - \varepsilon, 1 + \varepsilon ) \hat{A}_t \big] \right\} 
    ,
\end{split}
\label{eq:ppoloss_styled_simple}
\end{equation}
where $q$ represents query with updated information, $a$ is the corresponding ground-truth answer, $o$ denotes the sequence of generated tokens, $\hat{A}_t$ is the estimated advantage, and $\varepsilon$ is the clipping hyperparameter. The objective leverages the principle of importance sampling through the probability ratio $r_t(\theta)$ between the current policy ($\pi_{\theta}$) and the old policy ($\pi_{\theta_{\text{old}}}$):
\begin{equation}
    r_t(\theta) = \frac{\pi_{\theta}(o_t\mid q,o_{<t})}{\pi_{\theta_{\text{old}}}(o_t\mid q,o_{<t})} ,
    \label{eq:prob_ratio}
\end{equation}

\subsection{Stage-aware Reinforcement Learning}
\label{sec:stage_aware_rl}
To improve LLMs in complex multi-step reasoning tasks, we focus not only on the correctness of the final answer but also on optimizing the logical consistency and interpretability of the reasoning process itself. Traditional outcome-based rewards are often too sparse for multi-step reasoning, failing to provide effective guidance for the model's intermediate steps. For instance, a model might arrive at the correct answer through flawed reasoning (i.e., "lucky guess"), or an entire chain of reasoning could collapse due to a minor intermediate error. A sparse reward signal cannot effectively distinguish between these scenarios.
To address this challenge, we introduce a Stage-aware Reward mechanism. The core is a structured reward function designed to perform a fine-grained evaluation and provide incentives for the model's performance at different stages of the reasoning process.

\paragraph{Format Validation}
We first rigorously check whether the model's output adheres to the predefined tag structure, ensuring that tags such as \texttt{<think>}, \texttt{<decompose>}, \texttt{<action>}, and \texttt{<answer>} are correctly paired and appear in the proper sequence. Furthermore, we validate the internal structure within the <action> tag, verifying that sub-questions and their answers conform to the [Sub question X] and \texttt{\textbackslash boxed\{\}} formats. Any format violation resuls in a fixed penalty score (e.g., -1.0) and terminates further evaluation. This mechanism compels the model to learn to generate structured and parsable reasoning chains, which is fundamental for our fine-grained process assessment.

Once the output passes format validation, our reward function is composed of two primary components: a Process Score and an Outcome Score.

\paragraph{Process Score} This is the cornerstone of our stage-aware method, designed to assess the quality of the model's reasoning process. It is further broken down into three sub-components:

\textbf{(1) Hop Score:} It assesses whether model has correctly identified the number of reasoning "hops" required to solve the problem. We score this by verifying if the number of generated sub-questions matches the number of predefined reasoning steps. A correct reasoning framework begins with an accurate assessment of the problem complexity.

\textbf{(2) Decomposition Score:} This component evaluates the quality of the sub-questions formulated by the model. We employ a pre-trained Sentence Transformer to convert both the model-generated sub-questions and the ground-truth sub-questions into vector representations. The cosine similarity between these vectors is then calculated to measure their semantic equivalence. High-quality decomposition is a prerequisite for reaching the correct solution, and this score incentivizes the model to learn how to break down complex problems into a series of logically coherent and solvable sub-tasks.

\textbf{(3) Sub-answer Score:} This part measures model's ability to correctly solve each sub-question. We individually check the correctness of the answer provided for each sub-question (enclosed \texttt{boxed\{\}}). The score is proportional to the ratio of correct sub-answers. This provides direct feedback for each intermediate step, encouraging the model to maintain high accuracy throughout the entire reasoning chain.

\paragraph{Outcome Score} It evaluates the accuracy of the final answer. We extract the final answer generated by the model within the \texttt{<answer>} tags and compute F1 score against the ground truth. This ensures that the model's final output remains reliable.

\paragraph{Final Score} If format validation fails, the final score is -1; otherwise, the final score is a weighted combination of the process score ($R_{process}$) and the outcome score ($R_{outcome}$):
\begin{equation}
    R_{final} = 
    \alpha \cdot R_{process} + (1 - \alpha) \cdot R_{outcome}, 
\end{equation}
where $\alpha \in [0, 1]$ is a hyperparameter that controls the trade-off between these two components.

Through this stage-aware reward mechanism, the training signal is transformed from sparse and monolithic to dense and multi-dimensional. It informs the model not only \textit{if} it was correct, but \textit{which} intermediate steps were flawed. This fine-grained feedback significantly improves the model's ability to learn complex reasoning strategies, leading to more logical, interpretable, and robust reasoning processes.

\begin{table*}[t]
    \centering
    \resizebox{1.0\linewidth}{!}{%
    \begin{tabular}{lcccccccccc}
        \toprule
        \multirow{2}{*}{\bf Method}
            & \multicolumn{3}{c}{\textbf{2-hops}}
            & \multicolumn{3}{c}{\textbf{3-hops}}
            & \multicolumn{3}{c}{\textbf{4-hops}}
            & \multirow{2}{*}{\textit{\textbf{Avg.}}} \\
        \cmidrule(lr){2-4}\cmidrule(lr){5-7}\cmidrule(lr){8-10}
        & \textbf{w/o Distr.} & \textbf{w/ 2 Distr.} & \textbf{w/ 4 Distr.}
        & \textbf{w/o Distr.} & \textbf{w/ 2 Distr.} & \textbf{w/ 4 Distr.}
        & \textbf{w/o Distr.} & \textbf{w/ 2 Distr.} & \textbf{w/ 4 Distr.}
        & \\  
        \midrule
        ROME & \underline{12.00} & 12.00 & 11.99{\scriptsize\color{green}$\downarrow$}
            & \underline{8.83} & 8.95 & 9.11
            & \underline{5.46} & 5.68 & 5.50
            & 8.84 \\ 
        Mello & \underline{80.90} & 70.90{\scriptsize\color{red}$\downarrow$} & 65.80{\scriptsize\color{red}$\downarrow\downarrow$}
            & \underline{40.30} & 29.50{\scriptsize\color{red}$\downarrow$} & 30.40{\scriptsize\color{red}$\downarrow\downarrow$}
            & \underline{9.30} & 10.40 & 11.00
            & 38.72 \\  
        PokeMQA & \underline{84.10} & 77.80{\scriptsize\color{red}$\downarrow$} & 78.30{\scriptsize\color{green}$\downarrow$}
            & \underline{61.40} & 50.90{\scriptsize\color{red}$\downarrow$} & 49.40{\scriptsize\color{red}$\downarrow\downarrow$}
            & \underline{16.00} & 12.70{\scriptsize\color{green}$\downarrow$} & 9.10{\scriptsize\color{red}$\downarrow$}
            & 48.86 \\
            
        EditCoT & \underline{76.40} & 51.80{\scriptsize\color{red}$\downarrow\downarrow$} & 54.70{\scriptsize\color{red}$\downarrow\downarrow$}
            & \underline{44.00} & 16.10{\scriptsize\color{red}$\downarrow\downarrow$} & 16.90{\scriptsize\color{red}$\downarrow\downarrow$}
            & \underline{67.50} & 30.00{\scriptsize\color{red}$\downarrow\downarrow$} & 30.10{\scriptsize\color{red}$\downarrow\downarrow$}
            & 43.06 \\
        RAE & \underline{88.90} & 87.50{\scriptsize\color{green}$\downarrow$} & 85.30{\scriptsize\color{green}$\downarrow$}
            & \underline{71.10} & 60.10{\scriptsize\color{red}$\downarrow$} & 58.10{\scriptsize\color{red}$\downarrow\downarrow$}
            & \underline{76.30} & 65.50{\scriptsize\color{red}$\downarrow$} & 60.20{\scriptsize\color{red}$\downarrow\downarrow$}
            & 72.56 \\
        
        Reason-KE & \underline{97.00} & 96.70{\scriptsize\scriptsize\color{green}$\downarrow$} & 96.70{\scriptsize\color{green}$\downarrow$}
            & \underline{88.90} & 85.20{\scriptsize\color{green}$\downarrow$} & 84.80{\scriptsize\color{green}$\downarrow$}
            & \underline{95.60} & 85.80{\scriptsize\color{red}$\downarrow$} & 81.10{\scriptsize\color{red}$\downarrow$}
            & 90.20 \\       
        \cmidrule{1-11}
        Reason-KE++ & \underline{\textbf{98.90}} & \textbf{98.40}{\scriptsize\scriptsize\color{green}$\downarrow$} & \textbf{97.80}{\scriptsize\scriptsize\color{green}$\downarrow$} & \underline{\textbf{97.60}} & \textbf{95.30}{\scriptsize\scriptsize\color{green}$\downarrow$} & \textbf{94.30}{\scriptsize\scriptsize\color{green}$\downarrow$} & \underline{\textbf{98.80}} & \textbf{90.20}{\scriptsize\scriptsize\color{red}$\downarrow$} & \textbf{88.00}{\scriptsize\scriptsize\color{red}$\downarrow$} & \textbf{95.48} \\
        \bottomrule
    \end{tabular}}
    \caption{
   \textbf{ Multi-hop QA performance} is shown, with the best scores in \textbf{bold}. We compare the baseline (underlined, \textbf{no distractors}) against performance with \textbf{2 or 4 distractors}. The resulting performance change is categorized as:
    stable ({\scriptsize\color{green}$\downarrow$}, <6\% drop),
    significant ({\scriptsize\color{red}$\downarrow$}, >6\% drop), 
    or catastrophic ({\scriptsize\color{red}$\downarrow\downarrow$}, >12\% drop).
    }
    \label{table:hops}
\end{table*}

\begin{table*}[t]
    \centering
    \resizebox{1.0\linewidth}{!}{%
    \begin{tabular}{lcccccccccc}
        \toprule
        \multirow{2}{*}{\bf Method}
            & \multicolumn{3}{c}{\textbf{\#Edits: 1}}
            & \multicolumn{3}{c}{\textbf{\#Edits: 2}}
            & \multicolumn{3}{c}{\textbf{\#Edits: 3 \& 4}}
            & \multirow{2}{*}{\textit{\textbf{Avg.}}} \\
        \cmidrule(lr){2-4}\cmidrule(lr){5-7}\cmidrule(lr){8-10}
        & \textbf{w/o Distr.} & \textbf{w/ 2 Distr.} & \textbf{w/ 4 Distr.}
        & \textbf{w/o Distr.} & \textbf{w/ 2 Distr.} & \textbf{w/ 4 Distr.}
        & \textbf{w/o Distr.} & \textbf{w/ 2 Distr.} & \textbf{w/ 4 Distr.}
        & \\  
        \midrule
        ROME & \underline{9.36}  & 9.37  & 9.47 
            & \underline{9.81}  & 9.97  & 9.97
            & \underline{6.66}  & 6.85  & 6.70
            & 8.68 \\
        Mello & \underline{41.54}  & 35.41{\scriptsize\color{red}$\downarrow$} & 32.11{\scriptsize\color{red}$\downarrow$}
            & \underline{55.67}  & 48.83{\scriptsize\color{red}$\downarrow$} & 47.70{\scriptsize\color{red}$\downarrow$ }
            & \underline{30.60}  & 23.81{\scriptsize\color{red}$\downarrow$} & 25.24{\scriptsize\color{red}$\downarrow$}
            & 37.88 \\     
        PokeMQA & \underline{59.38}  & 54.35{\scriptsize\color{green}$\downarrow$} & 53.34{\scriptsize\color{red}$\downarrow$}
            & \underline{63.92}  & 56.23{\scriptsize\color{red}$\downarrow$} & 55.48{\scriptsize\color{red}$\downarrow$}
            & \underline{33.81}  & 26.19{\scriptsize\color{red}$\downarrow$} & 22.98{\scriptsize\color{red}$\downarrow$}
            & 47.30 \\     
        EditCoT & \underline{64.59} & 49.86{\scriptsize\color{red}$\downarrow\downarrow$} & 49.13{\scriptsize\color{red}$\downarrow\downarrow$}
            & \underline{64.57} & 35.52{\scriptsize\color{red}$\downarrow\downarrow$} & 39.46{\scriptsize\color{red}$\downarrow\downarrow$}
            & \underline{57.62} & 6.55{\scriptsize\color{red}$\downarrow\downarrow$} & 7.02{\scriptsize\color{red}$\downarrow\downarrow$}
            & 41.59 \\
        RAE & \underline{65.97} & 63.04{\scriptsize\color{green}$\downarrow$} & 60.11{\scriptsize\color{green}$\downarrow$}
            & \underline{81.07} & 68.98{\scriptsize\color{red}$\downarrow\downarrow$} & 67.39{\scriptsize\color{red}$\downarrow\downarrow$}
            & \underline{92.50} & 84.05{\scriptsize\color{red}$\downarrow$} & 78.57{\scriptsize\color{red}$\downarrow\downarrow$}
            & 73.52 \\  
        Reason-KE & \underline{89.84} & 84.08{\scriptsize\color{green}$\downarrow$} & 84.26{\scriptsize\color{green}$\downarrow$}
            & \underline{97.00} & 90.25{\scriptsize\color{red}$\downarrow$} & 85.85{\scriptsize\color{red}$\downarrow$}
            & \underline{95.00} & 94.64{\scriptsize\color{green}$\downarrow$} & 93.93{ \scriptsize\color{green}$\downarrow$}
            & 90.54 \\
        \cmidrule{1-11}
        Reason-KE++ &  \underline{\textbf{98.44}} & \textbf{91.49}{\scriptsize\color{red}$\downarrow$} & \textbf{90.12}{\scriptsize\color{red}$\downarrow$} &  \underline{\textbf{97.47}} & \textbf{95.13}{\scriptsize\color{green}$\downarrow$} & \textbf{93.44}{\scriptsize\color{green}$\downarrow$} &  \underline{\textbf{99.64}} & \textbf{98.10}{\scriptsize\color{green}$\downarrow$} & \textbf{97.50}{\scriptsize\color{green}$\downarrow$} & \textbf{95.70} \\
        \bottomrule
    \end{tabular}}
    \caption{
    \textbf{Multi-edit performance} is presented, with the best results in \textbf{bold} and all markers retaining the same meaning as in Table~\ref{table:hops}.
    }
    \label{table:edits}
    \vspace{-10pt}
\end{table*}

\section{Experiments}
\subsection{Experimental Setup}
\paragraph{Baselines and Models.}
We evaluate our approach against a parameter modification method (ROME~\cite{rome}) and several parameter preservation methods (MeLLo~\cite{mquake}, PokeMQA~\cite{pokemqa}, EditCoT~\cite{EDITCOT}, and RAE~\cite{rae}).
Most baselines and our method are implemented on Qwen2.5-instruct-7B~\cite{qwen25}.
For the training-free methods MeLLo and PokeMQA, preliminary experiments revealed suboptimal performance on the Qwen architecture; we thus evaluated them using the more powerful Deepseek-v3 API~\cite{deepseekv3} to ensure a fair assessment of their full capabilities.
To demonstrate generalizability, we  report performance on Llama3-8B-Instruct~\cite{grattafiori2024llama}. Further details for all baselines are in Appendix~\ref{sec:details_baselines}.

\paragraph{Datasets and Metrics.}
Our evaluation leverages two distinct benchmarks: the MQuAKE dataset~\cite{mquake}, designed for multi-hop QA in knowledge editing, and the DUNE dataset~\cite{dune}, which focuses on generalized editing.
For testing, we utilize the MQUAKE-CF-3k set (3,000 instances) and the Arithmetic, New-Info, and Scientific subsets from DUNE.
For our RL training, we use the MQUAKE-CF set, which has no data overlap with our test set.
Consistent with prior work~\cite{mquake}, we adopt \textit{Multihop-Accuracy} (measured by Exact Match, EM) as the primary metric.
Further details are in Appendix~\ref{sec:appendix_Datasets}.

\begin{table}[t]
    \centering
    \resizebox*{1.0\linewidth}{!}{%
    \begin{tabular}{lcccc}
        \toprule
        \multirow{2}{*}{\bf Method} 
            & \multicolumn{3}{c}{\textbf{Multi-hop acc}} 
            & \multirow{2}{*}{\textit{\textbf{Avg.}}} \\
        \cmidrule(lr){2-4}
        & \textbf{w/o Distr.} & \textbf{w/ 2 Distr.} & \textbf{w/ 4 Distr.} & \multicolumn{1}{c}{} \\
        \midrule
        EditCoT   & 51.26 & 23.13{\scriptsize\color{red}$\downarrow\downarrow$} & 24.20{\scriptsize\color{red}$\downarrow\downarrow$} & \underline{32.87} \\
        RAE   & 85.23 & 80.73{\scriptsize\color{green}$\downarrow$} & 78.96{\scriptsize\color{red}$\downarrow$} & \underline{81.64} \\
        Reason-KE   & 94.37 & 89.47{\scriptsize\color{green}$\downarrow$} & 87.53{\scriptsize\color{red}$\downarrow$} & \underline{90.46} \\
        \cmidrule{1-5} 
        Reason-KE++ & \textbf{95.86} & \textbf{92.37}{\scriptsize\color{green}$\downarrow$} & \textbf{91.00}{\scriptsize\color{green}$\downarrow$} & \underline{\textbf{93.08}} \\
        \midrule
    \end{tabular}}
    \caption{
    \textbf{Performance of \texttt{Llama-3-8B-Instruct} on MQuAKE-CF-3k}, presented using the same notational conventions as in Table~\ref{table:hops}.
    }
    \label{table:llama}
    \vspace{-10pt}
\end{table}

\begin{table}[t]
    \centering
    \resizebox*{1.0\linewidth}{!}{%
    \begin{tabular}{llcccc}
        \toprule
        \multirow{2}{*}{\bf Subest} 
        & \multirow{2}{*}{\bf Method} 
            & \multicolumn{3}{c}{\textbf{Acc}} 
            & \multirow{2}{*}{\textit{\textbf{Avg.}}} \\
        \cmidrule(lr){3-5}
        & & \textbf{w/o Distr.} & \textbf{w/ 2 Distr.} & \textbf{w/ 4 Distr.} & \multicolumn{1}{c}{} \\
        \midrule
        \multirow{2}{*}{Arithmetic}         
        & EditCoT & 92.30 & 89.20{\scriptsize\color{green}$\downarrow$} & 90.42{\scriptsize\color{green}$\downarrow$} & \underline{90.64} \\
        & Reason-KE & 97.46 & 95.11{\scriptsize\color{green}$\downarrow$} & 95.21{\scriptsize\color{green}$\downarrow$} & \underline{95.93} \\
        & Reason-KE++ & \textbf{98.03} & \textbf{97.84}{\scriptsize\color{green}$\downarrow$} & \textbf{99.15} & \textbf{98.34} \\
        \cmidrule(lr){1-6}
        \multirow{2}{*}{New-Info} 
        & EditCoT & 81.20 & 80.10{\scriptsize\color{green}$\downarrow$} & 78.30{\scriptsize\color{green}$\downarrow$} & \underline{79.87} \\
        & Reason-KE & 84.44 & 83.35{\scriptsize\color{green}$\downarrow$} & 84.06{\scriptsize\color{green}$\downarrow$} & \underline{83.95} \\
        & Reason-KE++ & \textbf{90.90 }& \textbf{90.90 }& \textbf{91.30} & \textbf{91.03} \\
        \cmidrule(lr){1-6}
        \multirow{2}{*}{Scientific}         
        & EditCoT & 81.03 & 81.23{\scriptsize\color{green}$\downarrow$} & 80.70{\scriptsize\color{green}$\downarrow$} & \underline{80.99} \\
        & Reason-KE & 82.31 & 80.71{\scriptsize\color{green}$\downarrow$} & 80.59{\scriptsize\color{green}$\downarrow$} & \underline{81.20} \\
        & Reason-KE++ & \textbf{91.71} & \textbf{91.45}{\scriptsize\color{green}$\downarrow$} &\textbf{ 92.50} & \textbf{91.89} \\
        \bottomrule
    \end{tabular}}
    \caption{
    \textbf{Performance on the subsets of the DUNE dataset.} All symbols adhere to the same conventions as detailed in Table~\ref{table:hops}.
    }
    \label{tab:dune_performance}
    \vspace{-10pt}
\end{table}

\paragraph{Distractors Selection.}
To systematically assess robustness, we introduce distractor facts into the evidence set $\mathcal{E}$.
Specifically, for each of the $m$ supporting facts required by a question, we add $k$ distractors, where $k \in \{0, 1, 2\}$ represents the interference level.
This amounts to a total of $n = m \times k$ distractor facts added per question.
Further details are in Appendix~\ref{sec:selection}.

\begin{table}[ht]
    \centering
    \resizebox{1.0\linewidth}{!}{%
    \begin{tabular}{lccc}
        \toprule
        \multirow{2}{*}{\bf Method}
            & \multicolumn{3}{c}{\textbf{Answer w/ exposed}} \\
        \cmidrule(lr){2-4}
        & \textbf{w/o Distr.} & \textbf{w/ 2 Distr.} & \textbf{w/ 4 Distr.} \\
        \midrule
        Mello & \underline{56.75} 
            & 48.06 ({\scriptsize \textcolor{red}{$\downarrow$8.69}}) 
            & 46.54 ({\scriptsize \textcolor{red}{$\downarrow$10.2}}) \\
        PokeMQA & \underline{60.21}
            & 51.30 ({\scriptsize \textcolor{red}{$\downarrow$8.91}}) 
            & 50.27 ({\scriptsize \textcolor{red}{$\downarrow$9.94}}) \\
        EditCoT & \underline{64.25} 
            & 25.49 ({\scriptsize \textcolor{red}{$\downarrow$38.8}}) 
            & 27.32 ({\scriptsize \textcolor{red}{$\downarrow$36.9}}) \\
        RAE & \underline{94.98} 
            & 88.82 ({\scriptsize \textcolor{red}{$\downarrow$6.16}}) 
            & 85.15 ({\scriptsize \textcolor{red}{$\downarrow$9.83}}) \\
        Reason-KE & \underline{97.08} 
            & 96.70 ({\scriptsize \textcolor{green}{$\downarrow$0.38}}) 
            & 96.71 ({\scriptsize \textcolor{green}{$\downarrow$0.37}}) \\
        \cmidrule{1-4}
        Reason-KE++ & \textbf{99.62} & \textbf{98.70}({\scriptsize \textcolor{green}{$\downarrow$0.92}}) & \textbf{98.27}({\scriptsize \textcolor{green}{$\downarrow$1.35}}) \\
        \bottomrule
    \end{tabular}}
    \caption{
    \textbf{Performance under the answer-exposed condition}, where {\scriptsize \textcolor{red}{$\downarrow$}} marks a significant degradation (>5\%) compared to the 'w/o Distr.' case, while {\scriptsize \textcolor{green}{$\downarrow$}} denotes a stable performance with a minimal drop (<1.5\%).
    }
    \label{table:expose_ablate}
    \vspace{-10pt}
\end{table}

\subsection{Main Results}
The comparative performance is detailed in Tables~\ref{table:hops} and \ref{table:edits}.
Our main findings are as follows:

\paragraph{Reason-KE++ consistently outperforms all other methods, especially in high-complexity scenarios.}
As shown in Tables~\ref{table:hops} and \ref{table:edits}, complex scenarios requiring deep reasoning (multi-hop) or dense information navigation (multi-edit) cause a sharp performance fall-off for most baselines.
Although Reason-KE's explicit chains are strong, its SFT-based process is not fully optimized and shows performance degradation in demanding settings (e.g., 4-hop with distractors).
Reason-KE++ addresses this by using reinforcement learning to optimize its structured reasoning template, achieving a more effective and faithful process.
This yields substantial gains, outperforming Reason-KE by approximately 5\% on average in both multi-hop and multi-edit settings and establishing new state-of-the-art results.

\paragraph{Reason-KE++ demonstrates superior robustness to irrelevant information.}
As shown in Tables~\ref{table:hops} and \ref{table:edits}, introducing distractor facts causes significant performance degradation for most baselines. Methods like MeLLo and EditCoT are particularly vulnerable, often experiencing catastrophic accuracy drops (marked by {\scriptsize\color{red}↓↓}) of over 12\%. Even RAE, which employs filtering, suffers a noticeable decline.
In stark contrast, both Reason-KE and Reason-KE++ maintain exceptionally stable performance (marked by {\scriptsize\color{green}↓}). This highlights that the explicit reasoning chain structure provides a strong defense, which our RL framework successfully maintains and enhances, effectively immunizing the model against noise.

\paragraph{Reason-KE++ Demonstrates Broad Generalizability and Superiority.}
To affirm our method's broad applicability, we evaluate it on a different LLM and a more diverse dataset.
First, when deployed on Llama-3-8B-Instruct (Table~\ref{table:llama}), Reason-KE++ again establishes itself as the top-performing method, achieving a 2.6\% average gain over Reason-KE, confirming its architectural advantages are model-agnostic.
Furthermore, on the DUNE dataset (Table~\ref{tab:dune_performance}), Reason-KE++ delivers substantial improvements across all categories, especially on the complex "New-Info" (+7.1\%) and "Scientific" (+10.7\%) subsets.
This demonstrates that the enhanced reasoning process is highly effective for diverse, open-ended editing tasks.

\paragraph{Reason-KE++ is Robust Against Answer Leakage.}
\label{sec:leakage}
Our analysis revealed a potential data artifact in the MQUAKE-CF-3K dataset: in 1,852 instances, the object $o^*$ of a supporting fact $(s, r, o^*)$ directly coincides with the final multi-hop answer $o^*_n$.
This `answer leakage' raises a critical concern that models might learn a shortcut (i.e., answer extraction) rather than performing genuine reasoning.
To test this, we created an 'answer-exposed' setting using only these instances and introduced distractors.
As shown in Table~\ref{table:expose_ablate}, while most baselines suffer substantial degradation—with EditCoT plummeting by over 36\%—revealing their heavy dependence on this shortcut, our methods demonstrate remarkable resilience.
Both Reason-KE and Reason-KE++ maintain near-perfect stability, with performance drops of less than 1.5\% (indicated by {\scriptsize \textcolor{green}{$\downarrow$}}).
This confirms our framework promotes a true reasoning process, effectively ignoring superficial cues from answer leakage.

\section{Analysis}

\begin{table*}[t]
    \centering
    \resizebox{0.8\linewidth}{!}{
    \begin{tabular}{lcccccc}
        \toprule
        \bf Method
            & \bf Multi-hop acc 
            & \bf Format acc 
            & \bf Hops acc 
            & \bf Sub acc
            & \bf Similarity
            & \textit{\bf Avg.} \\
        \midrule
        SFT & 81.90 & 84.93 & 21.13 & 14.13 & 51.74 & 50.77 \\
        \midrule
        + Outcome Score & 94.72 & 73.50 & 19.00 & 16.33 & 54.73 & 51.66 \\
        ++ Format Validation & 95.56 & 99.77 & 39.83 & 29.67 & 55.80 & 64.13 \\
        +++ Hop Score & 95.43 & 99.97 & 90.17 & 78.23 & 62.25 & 85.21 \\
        ++++ Sub-answer Score & \textbf{95.64} & 99.98 & 94.67 & \textbf{87.30} & 62.37 & 87.99 \\
        +++++ Dec. Score & 95.48 & \textbf{100.00} & \textbf{94.93} & 87.17 & \textbf{81.10} & \textbf{91.74} \\
        \bottomrule
    \end{tabular}}
    \caption{
        \textbf{Ablation study results, averaged across distractor levels.} 
        Each metric is the average performance over settings with 0, 2, and 4 distractors. Best results in each column are \textbf{bolded}. The metrics are defined as follows:
        \textbf{Multi-hop acc} is the accuracy of the final answer (EM);
        \textbf{Format acc} is the percentage of outputs adhering to the predefined format;
        \textbf{Hops acc} is the accuracy of the number of decomposed sub-questions matching the ground truth;
        \textbf{Sub acc} is the accuracy of intermediate \texttt{\textbackslash boxed\{\}} answers;
        \textbf{Similarity} is the semantic similarity score between generated and ground truth sub-questions.
    }
\label{table:ablation_summary_bold}
\end{table*}

\begin{figure*}[!t]
    \centering
\includegraphics[width=0.8\linewidth]{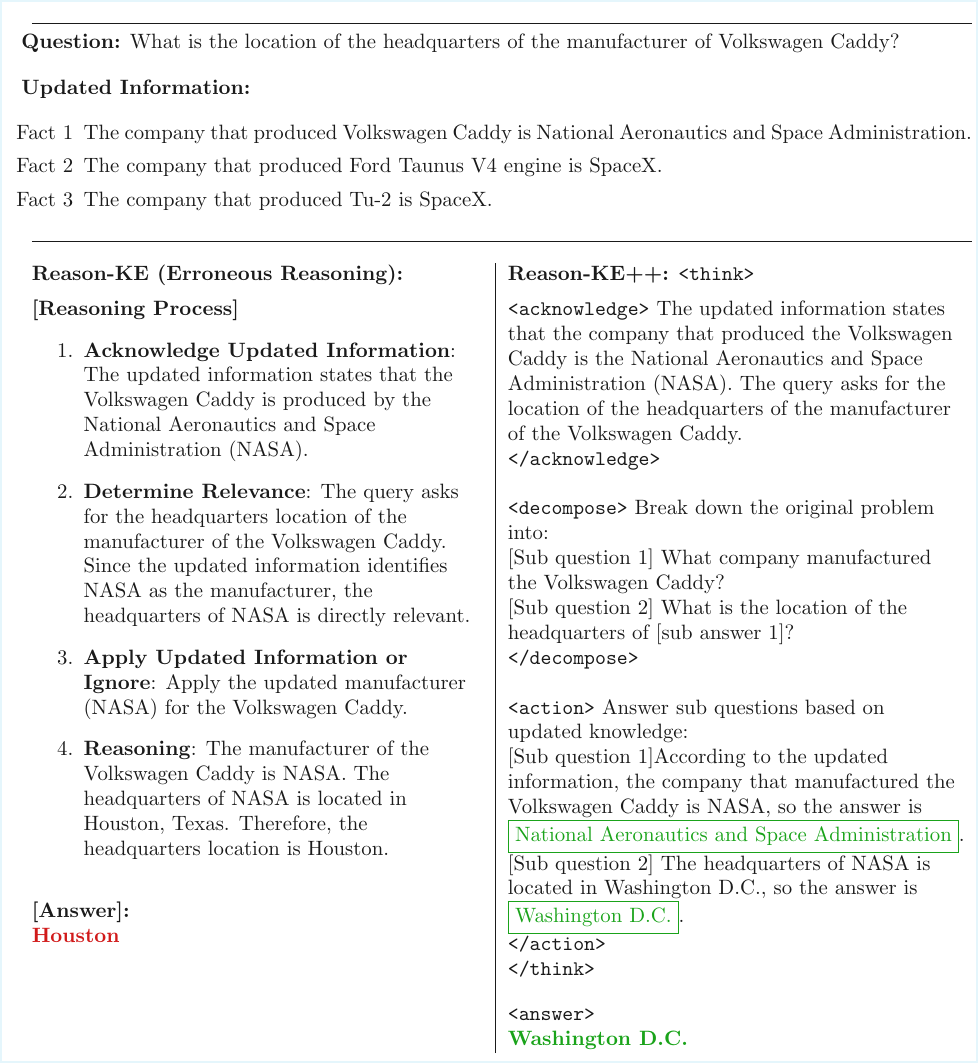}
    \caption{\textbf{Case study comparing Reason-KE~\cite{reasonke} and Reason-KE++}. `Reason-KE' defaults to its parametric prior (NASA $\rightarrow$ Houston), exhibiting factual hallucination. `Reason-KE++' uses structured decomposition to faithfully arrive at the correct answer.}
    \label{fig:case}
    
\end{figure*}

\subsection{Ablation Study of Reason-KE++}
To analyze the contribution of each component within our framework, we conducted a comprehensive ablation study.
As presented in Table~\ref{table:ablation_summary_bold}, a model trained via standard SFT acquires rudimentary reasoning skills but performs inadequately, highlighting the need for enhancement.
We thus initiated reinforcement learning, starting with a naive, outcome-only reward (`+ Outcome Score'). While this approach yields a superficial boost in `Multi-hop acc' (to 94.72), it proves to be a \textbf{deceptive trap}: the model fails to adhere to the desired format (`Format acc' drops to 73.50) and exhibits poor reasoning decomposition (`Hops acc' collapses to 19.00). This indicates the model has learned a new shortcut rather than a robust strategy.

To rectify this, we systematically integrated our process-oriented rewards. The introduction of each component demonstrably improves the quality of the reasoning process: `++ Format Validation' fixes structural integrity, `+++ Hop Score' dramatically enhances problem decomposition, and subsequent rewards refine the intermediate steps.
This step-wise refinement culminates in our final `Reason-KE++' model, which achieves holistic excellence across all metrics. This ablation validates that \textit{supervising the entire reasoning process, not just the outcome, is indispensable for building a powerful and faithful reasoner}.

\subsection{Case Study}

We present a case study in Figure~\ref{fig:case} to illustrate the fundamental difference in faithfulness between `Reason-KE'~\cite{reasonke} and `Reason-KE++'. When presented with a query and updated facts, both models correctly identify the relevant information (that NASA produced the Caddy) while discarding irrelevant distractors.

However, a critical divergence emerges in the subsequent reasoning. `Reason-KE' defaults to its widely known (but in this context, incorrect) parametric prior, stating NASA's headquarters is in "Houston". This leads to an erroneous final answer, demonstrating a clear vulnerability to \textbf{factual hallucination}.
In contrast, `Reason-KE++' demonstrates superior reasoning capability. Its explicit `<decompose>` step breaks the complex query into simpler, verifiable sub-problems. This structured approach forces the model to perform more fine-grained reasoning, thereby correctly identifying "Washington D.C." as NASA's headquarters.
This case illustrates that the \textit{structured reasoning framework of `Reason-KE++' is not merely a formatting preference but a crucial mechanism for ensuring step-by-step accuracy and mitigating error propagation, ultimately cultivating a more robust and trustworthy reasoner}.

\section{Related Work}
\label{sec:related_work}

\paragraph{Knowledge Editing}
Knowledge Editing (KE) aims to efficiently update factual knowledge in LLMs, with methods broadly categorized as parametric and non-parametric~\cite{easyedit,rome,mquake,wu-etal-2025-edit}. Parametric approaches, e.g., ROME~\cite{rome} and MEMIT~\cite{memit}, directly modify model weights by locating specific knowledge representations within FFN layers.
In contrast, non-parametric methods leverage in-context learning (ICL). Mello~\cite{mquake} pioneered the use of in-context editing (ICE), which decomposes complex queries into simpler sub-tasks and applies fine-grained edits through carefully crafted prompts. Building upon this foundation, subsequent works like PokeMQA~\cite{pokemqa} and EditCoT~\cite{EDITCOT} improved the robustness of this iterative approach.
More recently, Reason-KE~\cite{reasonke} departed from iterative ICL, employing Supervised Fine-Tuning (SFT) to generate an explicit, single-pass reasoning chain to solve the task. While this SFT approach established a new SOTA in robustness, it also introduced the "faithfulness gap" (see Figure 1) that our current work addresses.

\paragraph{Process-aware LLM Reinforcement Learning}
Reinforcement learning (RL) is pivotal for aligning LLMs with human preferences, as seen in methods like RLHF~\citep{ouyang2022training} and DPO~\citep{rafailov2023direct}. Beyond alignment, recent research increasingly applies RL to enhance specific capabilities, particularly complex reasoning. This has been demonstrated in both large-scale models like DeepSeek-R1~\citep{guo2025deepseek} and smaller, resource-efficient models~\citep{zeng2025simplerl, ye2025limo}.
A significant trend is the development of more efficient RL algorithms, such as GRPO~\citep{shao2024deepseekmath} and DAPO~\citep{yu2025dapo}, which make large-scale training on reasoning tasks more practical.

However, the application of RL to the niche, challenging field of \textit{multi-hop knowledge editing} remains largely unexplored. Our work addresses this critical gap. We demonstrate that naively applying RL with outcome-only rewards is a "deceptive trap" that collapses reasoning integrity (see Table 6). We are the first to show that to bridge the "faithfulness gap" left by SFT methods, a \textbf{Stage-aware Reward} mechanism is essential. Our approach pioneers the use of dense, process-level rewards to ensure LLMs can \textit{faithfully} reason over new, edited knowledge in complex scenarios.

\section{Conclusion}
\label{sec:conclusion}

We identified a critical ``faithfulness gap'' in SFT-based methods for multi-hop knowledge editing~\cite{reasonke, zhang2024uncovering}. These methods optimize for format mimicry, enabling LLM priors to override new facts and cause factual hallucinations. To solve this, we proposed \textbf{Reason-KE++}, an SFT+RL framework that instills process-level faithfulness. Its core is a \textbf{Stage-aware Reward mechanism} that provides dense supervision for intermediate reasoning steps, such as decomposition and sub-answer correctness. Crucially, we found naive outcome-only RL is a ``deceptive trap'' that collapses reasoning integrity (e.g., 19.00\% Hops acc). Our process-aware approach sets a new SOTA (95.48\%), proving that for complex tasks, aligning the \textit{process}, not just the \textit{outcome}, is essential for building trustworthy LLMs.

\normalem
\bibliography{acl_latex}

\appendix
\onecolumn

\section{Details of Dataset Construction}
\label{sec:Dataset_sample}
To construct the dataset for our cold-start phase, we began with the one-hop question-answering pairs from the COUNTERFACT dataset. For each pair, we generated a detailed, step-by-step reasoning process. Following a rigorous quality assurance protocol to filter out malformed or illogical samples, this procedure yielded a final, curated dataset of 718 instances.
To ensure consistency and quality, we guided the teacher model using a carefully designed prompt template, which is detailed below.

\begin{tcolorbox}
[colback=lightgray!20,colframe=darkgray!80,title= Reasoning Process Generation Prompt]
\label{tab:quality_prompt}
Please provide a reasoning process based on my following tasks and corresponding answers.
\newline
\newline
[Task]:Please acknowledge the updated information provided below and respond to the subsequent query. \newline
\newline
[Updated Information]: \newline
[Fact 1]Roblin Park is located in New South Wales. \newline
[Fact 2]The Eiffel Tower is located in London. \newline
\newline
[Query]:What is the capital city of the state where Roblin Park is located? \newline
\newline
<think>  \newline
<acknowledge>The updated information states that Roblin Park is located in New South Wales. And the query is "what is the capital city of the state where Roblin Park is located?"</acknowledge> \newline
\newline 
<decompose>Break down the original problem into: \newline
[Sub question 1]What state is Roblin Park located in? \newline
[Sub question 2]What is the capital of [sub answer 1]? </decompose> \newline
\newline
<action>Answer sub questions based on updated knowledge:\newline
[Sub question 1]Detected relevant to [Fact 1], so the answer is \texttt{\textbackslash boxed\{New South Wales\}}. \newline
[Sub question 2]No relevant facts were detected, but [sub answer 1] can be applied, so the answer is \texttt{\textbackslash boxed\{Sydney\}}.</action> \newline
</think> \newline
\newline
<answer>Sydney</answer>
\newline
\newline
[Task]:Please acknowledge the updated information provided below and respond to the subsequent query.
\newline
[Updated Information]: {\color{red}\texttt{<updated\_information>}}
\newline
[Query]: {\color{red}\texttt{<query>}}
\end{tcolorbox}

\section{Details of Experimental Setup}


\subsection{Details of Baselines}
\label{sec:details_baselines}
We evaluate ReasonKE++ against two main categories of knowledge editing techniques: a parameter modification method and several in-context editing approaches.
\paragraph{ROME~\cite{rome}}leverages causal mediation analysis to precisely locate and modify specific weights within a model's feed-forward networks. This update directly overwrites the stored factual knowledge. For our experiments, we implement ROME using the EasyEdit library~\cite{easyedit} with its default hyperparameter configuration.
\paragraph{MeLLo~\cite{mquake}}adopts a "plan-and-solve" methodology. It first deconstructs a complex query into simpler, solvable sub-questions, sequentially using retrieval to gather necessary information for each step. We follow the official implementation, adapting its prompts for instruction-tuned models and capping the retrieval process at four rounds.
\paragraph{PokeMQA~\cite{pokemqa}}refines the initial question decomposition stage. It prompts the LLM to generate a better-structured reasoning plan after augmenting the query with relevant knowledge. Our setup mirrors the official configuration, which includes a maximum of five interaction rounds and the use of their provided pre-trained Scope-Detector.
\paragraph{EditCoT~\cite{EDITCOT}}focuses on iteratively refining a model's reasoning trace. It starts by generating an initial Chain-of-Thought (CoT) based on the query. A specialized editor module then revises this CoT, integrating retrieved knowledge to correct inconsistencies or fill informational gaps. The model is subsequently prompted to generate the final answer based on this refined reasoning path. In line with the original work, we limit the maximum number of retrieval rounds to four.
\paragraph{RAE~\cite{rae}}externalizes knowledge into a graph structure. It trains the model to perform optimized retrieval and pruning over this knowledge graph, effectively navigating the graph to find the correct information needed to answer the query.
\paragraph{Reason-KE~\cite{reasonke}}was the first work to address the multi-hop knowledge editing problem by generating an explicit reasoning chain. However, Reason-KE was solely based on a standard Supervised Fine-Tuning (SFT) approach. While SFT can teach a model to mimic the format of a reasoning process, it lacks a dynamic mechanism to reward logical correctness or penalize unfaithful reasoning. This reliance on static examples makes the model can not developing true robust reasoning capability.

\subsection{Details of Datasets}
\label{sec:appendix_Datasets}
Table~\ref{datasetstatistic} provides a statistical breakdown of the MQUAKE-CF-3k dataset, which consists of 3,000 instances.
\begin{table}[h]
\centering
\begin{tabular}{llcccc}
\toprule
 \textbf{Datasets} & \textbf{\#Edits} & \multicolumn{1}{l}{\textbf{2-hop}} & \multicolumn{1}{l}{\textbf{3-hop}} & \multicolumn{1}{l}{\textbf{4-hop}} & \multicolumn{1}{l}{\textbf{Total}} \\ \hline
                                 & 1       & 513  & 356  & 224  & 1093 \\
                                 & 2       & 487  & 334  & 246  & 1067 \\
\multicolumn{1}{c}{MQUAKE-CF-3K} & 3       & -    & 310  & 262  & 572  \\
                                 & 4       & -    & -    & 268  & 268  \\
                                 & All     & 1000 & 1000 & 1000 & 3000 \\ 
\bottomrule
\end{tabular}
\caption{Statistics of MQuAKE-CF-3K datasets.}
\label{datasetstatistic}
\end{table}

\subsection{Implementation Details}
\label{sec:appendix_implementation}
We implemented our Reason-KE++ framework by trained two recent large language models: Llama3-8B-Instruct~\cite{grattafiori2024llama} and Qwen2.5-7B-Instruct~\cite{qwen25}. The training for each model followed our two-stage pipeline, encompassing both supervised fine-tuning (SFT) and reinforcement learning (RL). All experiments were conducted on a server equipped with 8 NVIDIA A100 (80GB) GPUs, and the entire training process required approximately 360 to 400 minutes per model. The specific hyperparameters used for training are detailed in Table~\ref{tab:hyperparameters}.

\begin{table}[!h]
\centering
{\begin{tabular}{lcc}
\toprule
\textbf{Hyperparameter} & \textbf{SFT} & \textbf{RL} \\ 
\midrule
Learning rate (Actor)   & 1e-5         & 1e-6        \\
Learning rate (Critic)  & -            & 1e-5        \\
Max sequence length     & 32768        & 1024        \\
Batch size              & 1            & 2048        \\
Optimizer               & AdamW        & AdamW       \\
Scheduler               & cosine       & -           \\
Weight decay            & 1e-4         & -           \\
Warmup ratio            & 0.05         & -           \\
KL coefficient          & -            & 0.001       \\
Training epochs         & 10           & 15          \\
\bottomrule
\end{tabular}}
\caption{\label{tab:hyperparameters}
\textbf{Hyper-parameters} for training Reason-KE++.}
\end{table}

\subsection{Details of Distractors Selection.}
\label{sec:selection}
We utilize Contriever~\cite{Contriever} to implement the TopK~\cite{toks} retrieval-based baseline\footnote{Note that better retrieval models, e.g., ReContriever~\cite{lei-etal-2023-unsupervised}, and exemplar selection methods~\cite{peng-etal-2024-revisiting} will improve the performance, but it is not the focus of this work.}. For each target fact requiring an edit, this method retrieves the top-k most similar post-edit examples from our dataset, where $ k \in{0,1,2}$.

\section{Used Scientific Artifacts}
\label{sec:Scientific}
Our work leverages several key open-source libraries to ensure reproducibility. We confirm that our use of these artifacts is in full compliance with their respective licenses and intended purposes.

\begin{itemize} 
    
    \item \textit{DeepSpeed (Apache-2.0 license)}~\footnote{ \url{https://github.com/deepspeedai/DeepSpeed}}: An optimization library used to enhance the efficiency and scale of large language model training.
    
    \item \textit{Transformers (Apache-2.0 license)}~\footnote{ \url{https://github.com/huggingface/transformers}},: The core framework providing the architectures and tools for the pre-trained language models used in NLP tasks.

    \item \textit{trl (Apache-2.0 license)}~\footnote{ \url{https://github.com/huggingface/trl}}: A specialized library employed to implement the Supervised Fine-tuning and reinforcement learning phase.

    \item \textit{verl (Apache-2.0 license)}~\footnote{ \url{https://github.com/volcengine/verl}}: A flexible, efficient, and production-ready RL training library for large language models (LLMs).

\end{itemize}

\end{document}